\documentclass[times,twocolumn,final]{elsarticle}
\usepackage{medima}
\usepackage{framed,multirow}
\usepackage{amssymb}
\usepackage{latexsym}
\usepackage{url}
\usepackage{xcolor}
\usepackage{hyperref}
\usepackage[cmex10]{amsmath}
\usepackage{amsthm}
\usepackage{mathrsfs}
\usepackage{graphicx}
\usepackage{float}
\usepackage{array}
\usepackage{epstopdf}
\usepackage{cite}
\usepackage{color}

\def\x{{\mathbf x}}

\definecolor{newcolor}{rgb}{.8,.349,.1}

\journal{Medical Image Analysis}

\begin{document}

\verso{S Minaee, R Kafieh, M Sonka, S Yazdani, G.J. Soufi}

\begin{frontmatter}

\title{Deep-COVID: Predicting COVID-19 From Chest X-Ray Images Using Deep Transfer Learning}

\author[1]{Shervin \snm{Minaee}\corref{cor1}}
\cortext[cor1]{Corresponding author: 
  Email: sminaee@snap.com}
\author[2]{Rahele \snm{Kafieh}\corref{cor2}}
\cortext[cor2]{Second Corresponding author}
%\fntext[fn1]{This is author footnote for second author.}
\author[3]{Milan \snm{Sonka}}
%% Third author's email
%\ead{author3@author.com}
\author[4]{Shakib \snm{Yazdani}}
\author[5]{Ghazaleh \snm{ Jamalipour Soufi}}

\address[1]{Snap Inc., Seattle, WA, USA}
\address[2]{Medical Image and Signal Processing Research Center, Isfahan University of Medical Sciences, Iran}
\address[3]{Iowa Institute for Biomedical Imaging, The University of Iowa, Iowa City, USA}
\address[4]{ECE Department, Isfahan University of Technology, Iran}
\address[5]{Radiology Department, Isfahan University of Medical Sciences, Isfahan, Iran}

\received{24 April 2020}
%\finalform{10 May 2013}
%\accepted{13 May 2013}
%\availableonline{15 May 2013}
%\communicated{S. Sarkar}

\begin{abstract}
The COVID-19 pandemic is causing a major outbreak in more than 150 countries around the world, having a severe impact on the health and life of many people globally. 
One of the crucial step in fighting  COVID-19 is the ability to detect the infected patients early enough, and put them under special care. 
Detecting this disease from radiography and radiology images is perhaps one of the fastest ways to diagnose the patients.
Some of the early studies showed specific  abnormalities in the chest radiograms of patients infected with COVID-19.
Inspired by earlier works, we study the application of deep learning models to detect COVID-19 patients from their chest radiography images.
We first prepare a dataset of 5,000 Chest X-rays from the publicly available datasets. Images exhibiting  COVID-19 disease presence were identified by board-certified radiologist.
Transfer learning on a subset of 2,000 radiograms was used to train four popular convolutional neural networks, including ResNet18, ResNet50, SqueezeNet, and DenseNet-121, to identify COVID-19 disease in the analyzed chest X-ray images.
We evaluated these models on the remaining 3,000 images, and most of these networks achieved a sensitivity rate of 98\% ($\pm$ 3\%), while having a specificity rate of around 90\%.
Besides sensitivity and specificity rates, we also present the receiver operating characteristic (ROC) curve, precision-recall curve, average prediction, and confusion matrix of each model.
We also used a technique to generate heatmaps of lung regions potentially infected by COVID-19 and  show that the generated heatmaps contain most of the infected areas annotated by our board certified radiologist.
While the achieved performance is very encouraging, further analysis is required on a larger set of COVID-19 images, to have a more reliable estimation of accuracy rates.
The dataset, model implementations (in PyTorch), and evaluations, are all made publicly available for research community at \url{https://github.com/shervinmin/DeepCovid.git}
\end{abstract}


\end{frontmatter}

%\linenumbers

\section{Introduction}
\label{intro}
Since December 2019, a novel corona-virus (SARS-CoV-2) has spread from Wuhan to the whole China, and many other countries. By April 18, more than 2 million confirmed cases, and more than 150,000 deaths were reported in the world \citep{Radio1}.
Due to unavailability of therapeutic treatment or vaccine for novel COVID-19 disease, early diagnosis is of real importance to provide the opportunity of immediate isolation of the suspected person and to decrease the chance of infection to healthy population. Reverse transcription polymerase chain reaction (RT-PCR) or gene sequencing for respiratory or blood specimens are introduced as main screening methods for COVID-19 \citep{Radio2}. However, total positive rate of RT-PCR for throat swab samples is reported to be 30 to 60 percent, which accordingly yields to un-diagnosed patients, which may contagiously infect a huge population of healthy people \citep{Radio3}. 
Chest radiography imaging (e.g., X-ray or computed tomography (CT) imaging) as a routine tool for pneumonia diagnosis is easy to perform with fast diagnosis. Chest CT has a high sensitivity for diagnosis of COVID-19 \citep{Radio4} and X-ray images show visual indexes correlated with COVID-19 \citep{Radio5}.
The reports of chest imaging demonstrated multilobar involvement and peripheral airspace opacities. The opacities most frequently reported are ground-glass (57\%) and mixed attenuation (29\%) \citep{Radio6}. 
During the early course of COVID-19, ground glass pattern is seen in areas that edges  the pulmonary vessels and may be difficult to appreciate visually \citep{Radio7}.  Asymmetric patchy or diffuse airspace opacities are also reported for COVID-19 \citep{Radio8}.  Such subtle abnormalities can only be interpreted by expert radiologists. Considering huge rate of suspected people and limited number of trained radiologists, automatic methods for identification of such subtle abnormalities can assist the diagnosis procedure and increase the rate of early diagnosis with high accuracy.  Artificial intelligence (AI)/machine learning solutions are potentially powerful tools for solving such problems.

\begin{figure}[t]
\begin{center}
    \includegraphics[width=0.99\linewidth]{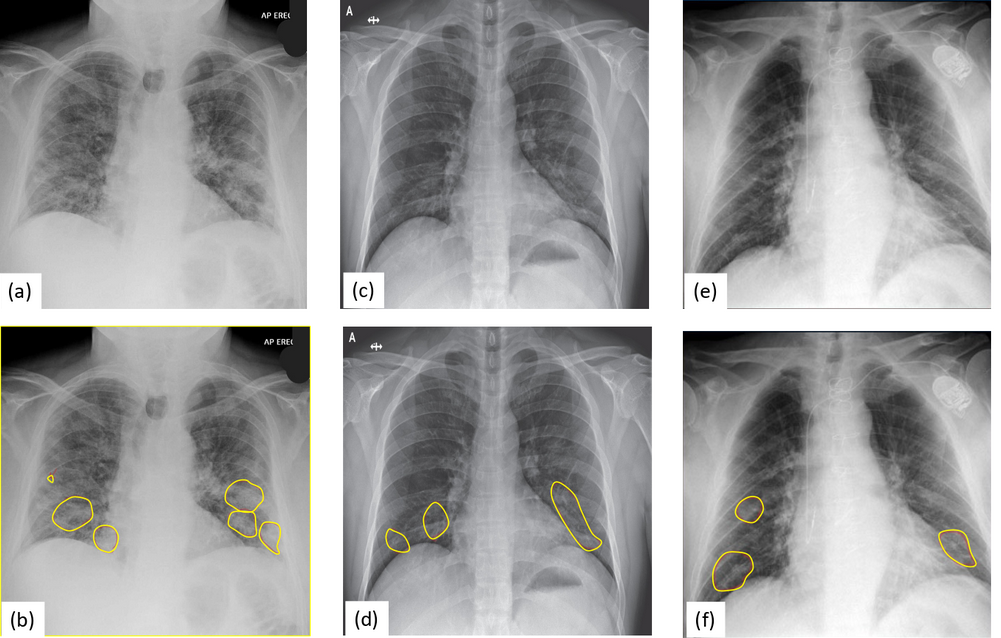}
\end{center}
  \caption{Three sample COVID-19 images, and the corresponding marked areas by our radiologist.
  %\xxx{MS: is it possible to make the radiologist markings better visible? Currently impossible to see unless you zoom-in a lot. Also, what is the difference between a marker and an outlined area?}
  }
  \label{fig:sample3}
\end{figure}

So far, due to the lack of availability of public images of COVID-19 patients,  detailed studies reporting solutions for automatic detection of COVID-19 from X-ray (or Chest CT) images are not available.
Recently a small dataset of COVID-19 X-ray images was collected, which made it possible for AI researchers to train machine learning models to perform automatic COVID-19 diagnostics from X-ray images \citep{covid_chest_paper}.
These images were extracted from academic publications reporting the results on COVID-19 X-ray and CT images.
With the help of a board-certified radiologist, we re-labeled those images, and only kept  ones a clear sign of COVID-19 as determined by our radiologist. 
Three sample images with their corresponding marked areas are shown in Figure \ref{fig:sample3}.
We then used a subset of images from ChexPert \citep{chexpert} dataset, as the negative samples for COVID-19 detection.
The combined dataset has around 5,000 Chest X-ray images (called COVID-Xray-5k), which is divided into 2,000 training, and 3,000 testing samples.
%It is worth mentioning that some of the earlier works in the past few weeks used the images from pediatric patients of one to five years old (from a Kaggle competition) as the negative class, which may not be the best idea, as there is a big difference among the age range of the positive and negative class in that case.
 
A machine a learning framework was employed to predict COVID-19 from Chest X-ray images. 
Unlike the classical approaches for medical image classification which follow a two-step procedure (hand-crafted feature extraction+recognition), we use an end-to-end deep learning framework which directly predicts the COVID-19 disease from raw images without any need of feature extraction.
%where in the first step a set of hand-crafted features (such as the mean or variance of image intensity in different regions, or distribution of edges) are extracted from images, and in the second step a classification algorithm (such as support vector machine (SVM), logistic regression, or decision trees) are used to perform recognition/classification. Although some of these approaches achieve accurate prediction, they involve a lot of pre-processing and use several hand-crafted features, which may not be optimal when going from one dataset to another one (collected under different conditions).
Deep learning based models (and more specifically convolutional neural networks (CNN)) have been shown to outperform the classical AI approaches in most of computer vision and and medical image analysis tasks in recent years, and have been used in a wide range of problems from classification, segmentation, face recognition, to super-resolution and image enhancement  \citep{cnn,cnn2,cnn3,cnn4,cnn5}.
%Convolutional neural networks have been used for various problems such as image classification, segmentation, super-resolution, image captioning, emotion recognition from images, face recognition, and object detection, and significantly improved the performance over traditional approaches \cite{cnn1,cnn2,cnn3,cnn4,cnn5,cnn6,cnn7,cnn8,cnn9}. More interestingly, it is shown that the features learned from some of these deep architectures can be transferred to other tasks easily, i.e. one can get the features from a trained model for a specific task and use it for a different task, by training a classifier/predictor on top of it \cite{offshelf}.

Here, we train 4 popular convolutional networks which have achieved promising results in several tasks during recent years (including ResNet18, ResNet50, SqueezeNet, and DenseNet-161) on COVID-Xray-5k dataset, and analyze their performance for COVID-19 detection. 
%Since, having a large-scale dataset (with several examples for each class of label) is crucial for successful training of a deep model from scratch, and so far we do not have many labeled images as COVID-19, we fine-tune well-known pre-trained models on this dataset, which can result in a well-performing model with a much fewer labeled samples. We also employ data augmentation techniques (such as flipping, small rotation, adding small amount of distortions) to increase the number of samples for each class.
Since so far there is a limited number of X-ray images publicly available for the COVID-19 class, we cannot simply train these models from scratch. Two strategies were adopted to address the COVID-19 image scarcity issue in this work:
\begin{itemize}
    \item We use data augmentation to create transformed version of COVID-19 images (such as flipping, small rotation, adding small amount of distortions), to increase the number of samples by a factor of 5.
    \item Instead of training these models from scratch, we fine-tune the last layer of the pre-trained version of these models on ImageNet. In this way, the model can be trained with less labeled samples from each class.
\end{itemize}
The above two strategies helped train these networks with the available images, and achieve reasonable performance on the test set of 3,000 images.
%The best performing model out of the above four networks, achieves a sensitivity of 98\%, and specificity of around 92\%. 
Since the number of samples for the COVID-19 class is limited, we also calculate the confidence interval of the performance metrics. 
To report a summarizing performance of these models, we  provide the Receiver operating characteristic (ROC) curve, and area under the curve (AUC) for each of these models.

%\xxx{MS: You list all these in the Conclusion - not sure you need this list here -- if removed from here, you can make the Conclusion section little more detailed. This is a minor point.}
Here are the main contributions of this paper:
\begin{itemize}
    \item We prepared a dataset of 5,000 images with binary labels, for COVID-19 detection from Chest X-ray images. This dataset can serve as a benchmark for the research community. The images in COVID-19 class, are labeled by a board-certified radiologist, and only those with a clear sign are used for testing purpose.
    \item We trained four promising deep learning models on this dataset, and evaluated their performance on a test set of 3,000 images. Our best performing model achieved a sensitivity rate of 98\%, while having a specificity of 92\%. 
    %\xxx{MS: I am not sure where are these numbers coming from - I do not see them in Tables 2-5, can you please check and maybe make it clear where they come from if correct?}
   % REPETITIVE \item We also provided the ROC curve, AUC, and the histogram of the predicted scores by these models.
    \item We provided a detailed experimental analysis on the performance of these models, in terms of sensitivity, specificity, ROC curve, area under the curve, precision-recall curve, and  histogram of the predicted scores.
    \item We provided the heatmaps  of the most likely regions, which are infected due to Covid-19, using a deep visualization technique.
    \item We made the dataset, the trained models, and the implementation publicly available.
\end{itemize}

It is worth to mention that while very encouraging, given the amount of the labeled data the result of this work is still preliminary and more concrete conclusion requires further experiments on a larger dataset of COVID-19 labeled X-ray images.
We believe this work can serve as a benchmark for future works and comparisons.

%\xxx{MS: This next paragraph is  not really needed since the paper structure is quite standard.}
The structure of the rest of this paper is as follows.
Section \ref{sec_dataset} provides a summary of the prepared \textbf{COVID-Xray-5k Dataset}.
Section \ref{sec_method} presents the description of the overall proposed framework. 
Section \ref{sec_result}  provides the experimental studies and comparison with previous works.
And finally the paper is concluded in Section \ref{sec_conc}.

\section{COVID-Xray-5k Dataset}
\label{sec_dataset}
Chest X-ray images from two datasets formed the COVID-Xray-5k dataset that contains 2,084 training and 3,100 test images.

One of the used datasets is the recently published \textbf{Covid-Chestxray-Dataset}, which contains a set of images from publications on COVID-19 topics, collected by Joseph Paul Cohen \citep{covid_chest, covid_chest_paper}. 
This dataset contains a mix of chest X-ray and CT images. 
As of May 3, 2020, it contained 
%\xxx{MS: this looks funny ``over 200 out of which 203'' -- I know that you have more than 200 but it still reads funny, do you have an exact number greater than 203 to put here?}
250 X-ray images of COVID-19 patients, from which 203 images are anterior-posterior view.
It is mentioned that this dataset is continuously updated.
It also contains some meta-data about each patients, such as sex and age.
Our COVID-19 images are all coming from this dataset. 
Based on our board-certified radiologist advice, only anterior-posterior images are kept for Covid-19 prediction, as the lateral images are not suitable for this purpose.
The anterior-posterior images were examined by our board-certified radiologist, and the ones without even the slightest radiographic signs of Covid-19 were removed from dataset. Out of 203 interior-exterior X-ray images of COVID-19, 19 of them were excluded, and 184 images (which showed clear signs of COVID-19) were kept by our radiologist. 
This way, we can  provide the community a more cleanly labeled dataset.
Out of these images, we chose 100 COVID-19 images to include in the test set (to meet some maximum confidence interval value), and 84 COVID-19 images for the training set. 
Data augmentation is  applied to the training set to increase the number of COVID-19 samples to 420 as described above.
We made sure all images for each patient go only to one of the training or test sets. 
It is worth mentioning that our radiologist marked the regions with specific signs of Covid-19.

Since the number of Non-Covid images was very small in the \citep{covid_chest} dataset, additional images were employed from the \textbf{ChexPert} dataset \citep{chexpert}, a large public dataset for chest radiograph interpretation consisting of 224,316 chest radiographs of 65,240 patients, labeled for the presence of 14 sub-categories (no-finding, Edema, Pneumonia, etc.). 
For the non-COVID samples in the training set, we only used images belonging to a single sub-category, composed of 700 images from the no-finding class and 100 images from each remaining 13 sub-classes, resulting in 2,000 non-COVID images. 

As for the Non-COVID samples in the test dataset, we selected 1,700 images from the no-finding category and around 100 images from each remaining 13 sub-classes in distinct sub-folders, resulting in 3000 images in total. 
%These images plus 40 COVID-19 images form \cite{covid_chest} have made the test part of COVID-Xray-5k dataset with more than 3k images. 
The exact number of images of each class for both training and testing is given in Table \ref{dataset_count}.

\begin{table}[ht]
\centering
  \caption{Number of images per category in COVID-Xray-5k dataset.}
  \centering
\begin{tabular}{|c|c|c|}
\hline
Split  & COVID-19 & Non-COVID\\
\hline
Training Set & 84 (420 after augmentation) & 2000  \\ \hline
Test Set  & 100 &  3000 \\ \hline
\end{tabular}
\label{dataset_count}
\end{table}

\begin{figure}[h]
\begin{center}
    \includegraphics[width=1.1\linewidth]{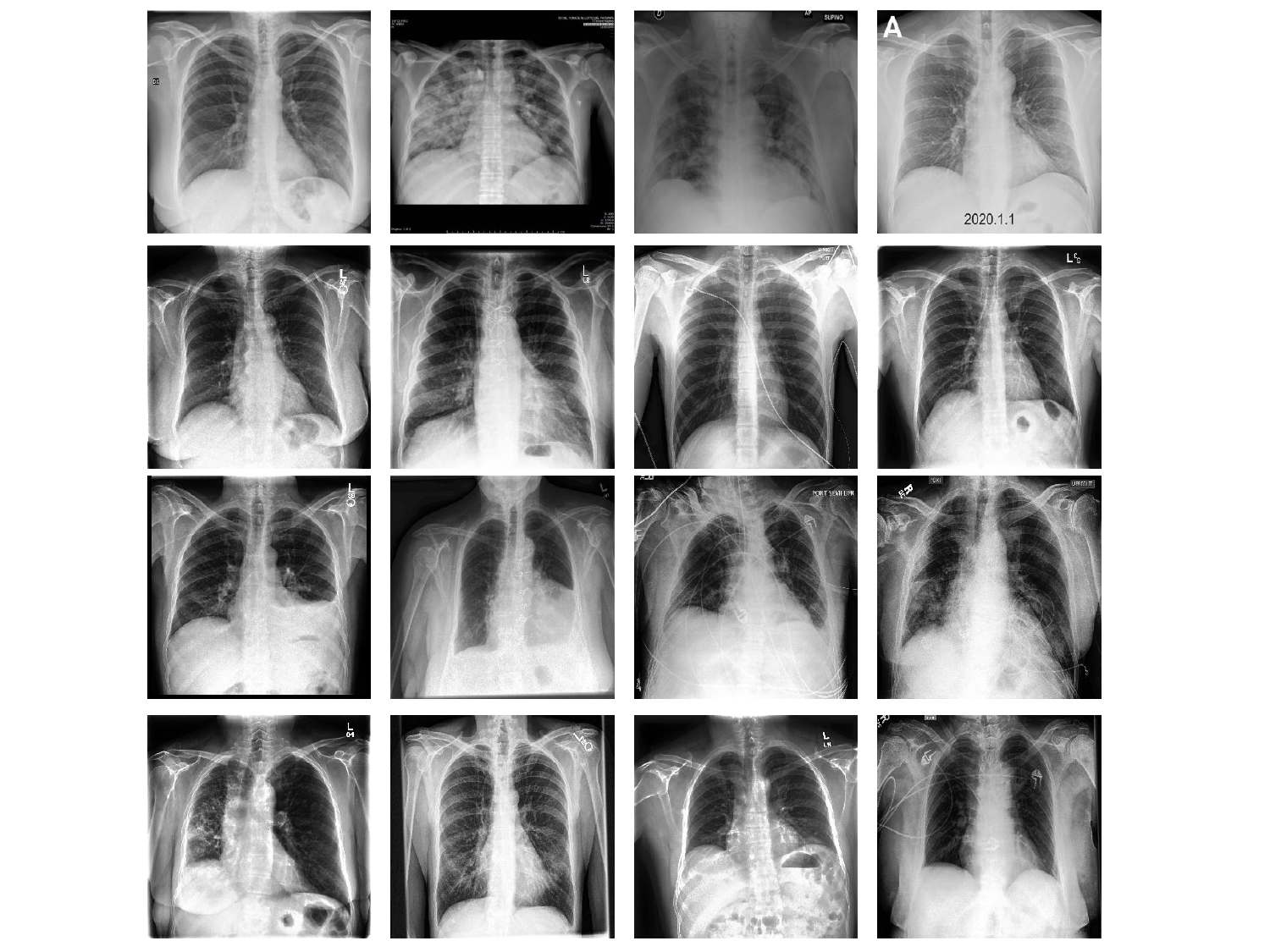}
\end{center}
  \caption{Sample images from COVID-Xray-5k dataset. The images in the first row show 4 COVID-19 images. The images in the second row are 4 sample images of no-finding category in Non-COVID images from \textbf{ChexPert}. The images in the third and fourth rows give 8 sample images from other sub-categotries in \textbf{ChexPert}.}
  \label{dataset_images}
\end{figure}

Figure \ref{dataset_images} shows 16 sample images from COVID-Xray-5k dataset, including 4 COVID-19 images (the first row), 4 normal images from ChexPert (the second row), and 8 images with one of the 13 diseases in ChexPert (third and fourth rows).

It is worth mentioning that, there is wide variation inn the resolution of images in this dataset. There are some low-resolution images in Covid-19 class (below 400x400), and some high resolutions ones (more than 1900x1400). 
This is a positive point for the models that can achieve a reasonable high accuracy on this dataset, despite this variable image resolution and imagery methodology. 
Collecting all images in a super-controlled environment that results in high-resolution and super-clean images, although desired, is not always doable, and as machine learning field progresses, more and more focus is directed toward models and frameworks that can work reasonably well on variable resolution, quality, and small-scale labeled datasets.
Also the images of Covid-19 class are collected from multiple sources by the original provider, and some of them may show a different dynamic range from other ones (and also from ChexPert), but during the training all images are normalized to the same distribution to make model less sensitive to that.

\section{The Proposed Framework}
\label{sec_method}
%Since so far, the number of publicly available images, which are labeled as COVID-19 are very limited, it may not be possible to train a deep convolutional neural network from scratch to detect COVID-19 from X-ray images.
To overcome the limited data sizes, 
%we use a well-known strategy in machine learning, called "
transfer learning was used to fine-tune four popular pre-trained deep neural networks on the training images of COVID-Xray-5k dataset.
%We will first provide a quick introduction of transfer learning, and then discuss the proposed framework. 
%By initializing the model weights as the pre-trained one on ImageNet, we will start from a network state which has a good understanding of general image description.

\subsection{Transfer Learning Approach}
In transfer learning, a model trained on one task is re-purposed to another related task, usually by some adaptation toward the new task.
For example, one can imagine using an image classification model trained on ImageNet (which contains millions of labeled images) to initiate task-specific learning for COVID-19 detection on a smaller dataset.
%\xxx{citation missing, and I think that ImageNet now has about 14M images.}
%It would be plausible to use the representation learned by a model, trained for general-purpose classification, for a different image processing task. There have been many works based on pre-trained deep learning models to perform a different task in the past few years. 
Transfer learning is mainly useful for tasks where enough training samples are not available to train a model from scratch, such as medical image classification for rare or emerging diseases.
%, in which sufficiently large numbers of labeled samples may not be available. 
This is especially the case for models based on deep neural networks, which have a large number of parameters to train.
By using transfer learning, the model parameters start with already-good initial values that only need some small modifications to be better curated toward the new task.
%Figure \ref{fig:VGG16} shows the block-diagram of a sample transfer learning approach for histology images based on a pre-trained CNN model.

There are two main ways in which the pre-trained model is used for a different task. 
In one approach, the pre-trained model is treated as a feature extractor (i.e., the internal weights of the pre-trained model are not adapted to the new task), and a classifier is trained on top of it to perform classification.  
In another approach, the whole network, or a subset thereof, is fine-tuned on the new task. Therefore the pre-trained model weights are treated as the initial values for the new task, and are updated during the training stage.

In our case, since the number of images in the COVID-19 category is very limited, we only fine-tune the last layer of the convolutional neural networks, and essentially use the pre-trained models as a feature extractor. 
We evaluate the performance of four popular pre-trained models, ResNet18 \citep{resnet}, ResNet50 \citep{resnet}, SqueezeNet \citep{squeeze}, and DenseNet-121 \citep{densenet}.
In the next section we provide a quick overview of the architecture of these models, and how they are used for COVID-19 recognition.

\subsection{COVID-19 Detection Using Residual ConvNet -- ResNet18 and ResNet50}
One of the models used in this work, is the pre-trained ResNet18, trained on ImageNet dataset.
ResNet is one of the most popular CNN architecture, which provides easier gradient flow for more efficient training, and was the winner of the 2015 ImageNet competition. 
The core idea of ResNet is introducing a so-called \textit{identity shortcut connection} that skips one or more layers. 
%as shown in Figure \ref{fig:Resnet}.
This would help the network to provide a direct path to the very early layers in the network, making the gradient updates for those layers much easier.
\iffalse
\begin{figure}[h]
\begin{center}
   \includegraphics[width=0.7\linewidth]{img/resnet.png}
\end{center}
   \caption{The residual block used in ResNet model, courtesy of \citep{resnet}}
\label{fig:Resnet}
\end{figure}
\fi

\begin{figure}[h]
\begin{center}
   \includegraphics[page=2,width=0.95\linewidth]{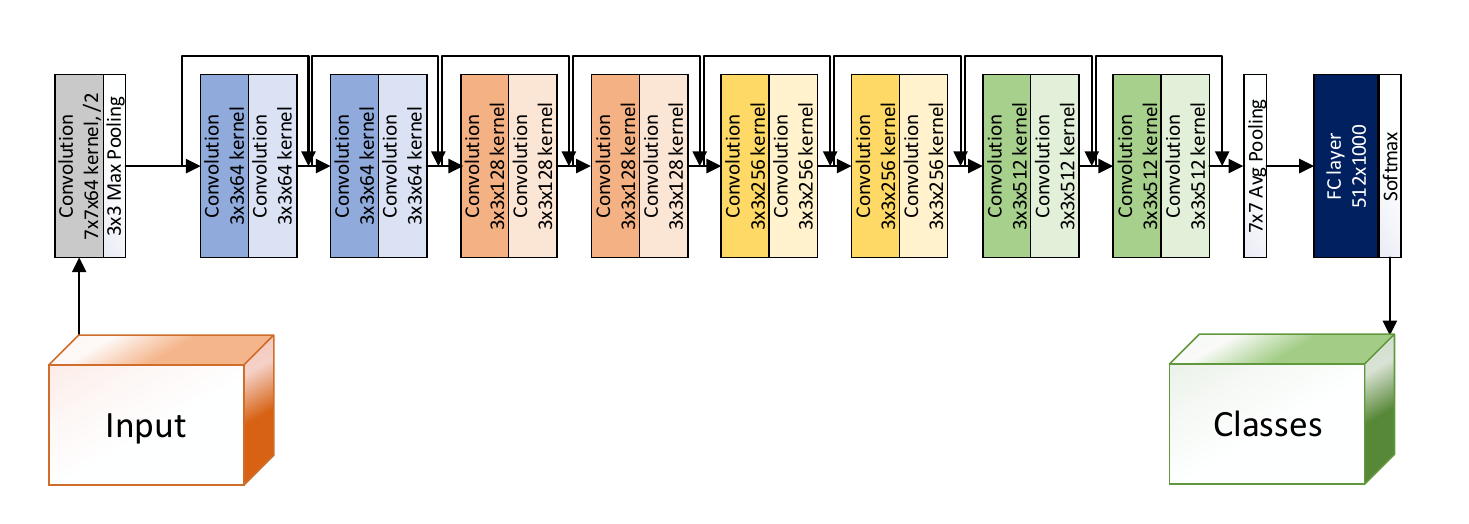}
\end{center}
   \caption{The architecture of ResNet18 model \citep{resnet}.}
\label{fig:ResnetNN}
\end{figure}

The overall block diagram of ResNet18 model, and how it is used for COVID-19 detection is illustrated in Figure \ref{fig:ResnetNN}.
ResNet50 architecture is pretty similar to ResNet18, the main difference being having more layers.

\subsection{COVID-19 Detection Using SqueezeNet}
\begin{figure}[h]
\begin{center}
   \includegraphics[page=2,width=0.95\linewidth]{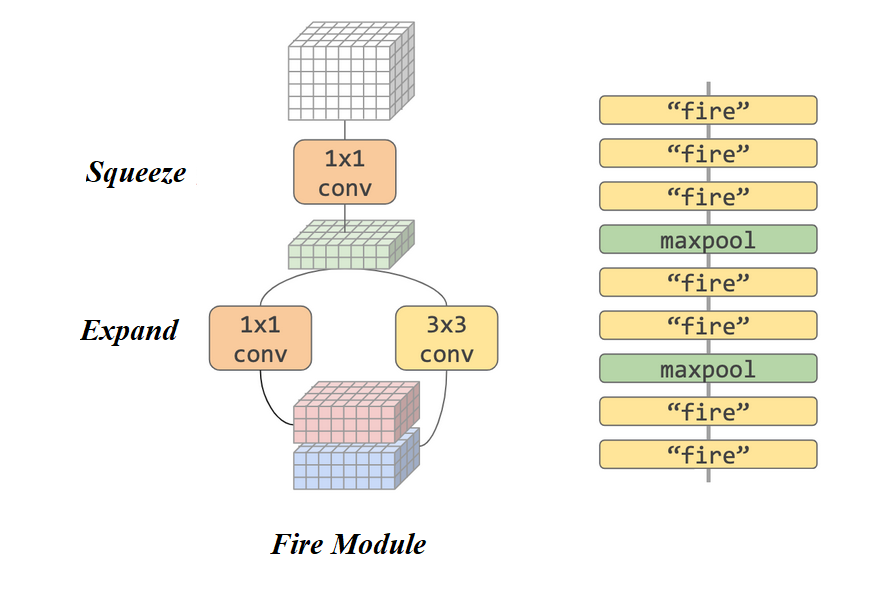}
\end{center}
   \caption{The architecture of SqueezeNet based on "fire modules". Courtesy of Google \citep{googlecolab}.}
\label{fig:squeezenet}
\end{figure}

SqueezeNet \citep{squeeze} proposed by Iandola et al., is a small CNN architecture, which achieves AlexNet-level \citep{alexnet} accuracy on ImageNet with 50x fewer parameters. 
Using model compression techniques, the authors were able to compress SqueezeNet to less than 0.5MB, which made it very popular for applications that require light-weight models.  
They alternate a 1x1 layer that "squeezes" the incoming data in the vertical dimension followed by two parallel 1x1 and 3x3 convolutional layers that "expand" the depth of the data again.
Three main strategies used in SqueezeNet includes: replace 3x3 filters with 1x1 filters, decrease the number of input channels to 3x3 filters, Down-sample late in the network so that convolution layers have large activation maps.
Figure \ref{fig:squeezenet} shows the architecture of a simple SqueezeNet.

\subsection{COVID-19 Detection Using DenseNet}
\begin{figure}[h]
\begin{center}
   \includegraphics[page=2,width=0.86\linewidth]{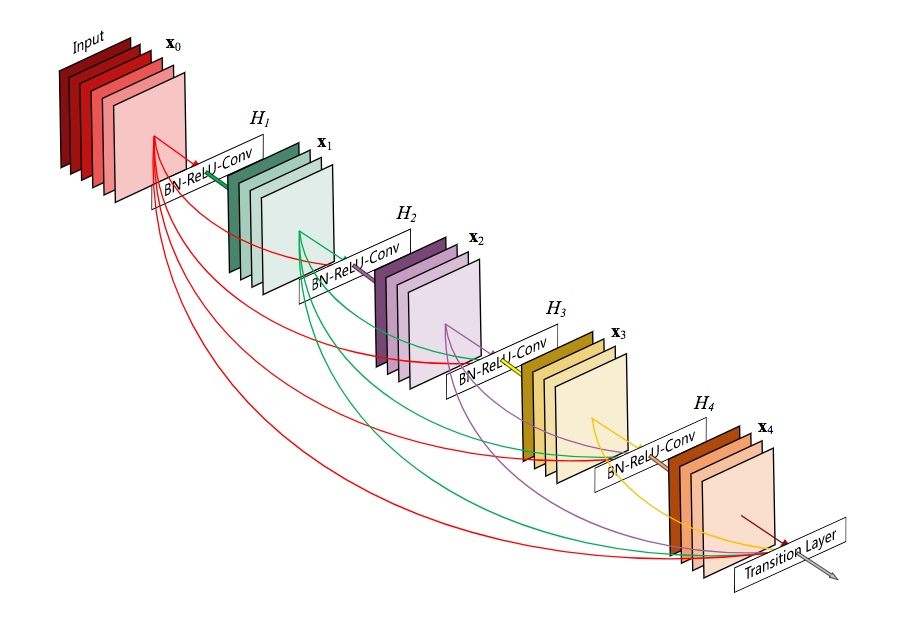}
\end{center}
   \caption{The architecture of a DenseNet with 5 layers, with expansion of 4. Courtesy of model \citep{densenet}.}
\label{fig:densenet}
\end{figure}

Dense Convolutional Network (DenseNet) is another popular architecture \citep{densenet}, which was the winner of the 2017 ImageNet competition. 
In DenseNet, each layer obtains additional inputs from all preceding layers and passes on its own feature-maps to all subsequent layers. 
Each layer is receiving a “collective knowledge” from all preceding layers.
Since each layer receives feature maps from all preceding layers, network can be thinner and compact, i.e., number of channels can be fewer (so, it have higher computational efficiency and memory efficiency). 
The architecture of sample DenseNet is shown in Figure \ref{fig:densenet}.

\subsection{Model Training}
All employed models are trained with a cross-entropy loss function, which tries to minimize the distance between the predicted probability scores, and the ground truth probabilities (derived from labels), and is defined as:
%To reduce the chance of over-fitting the $\ell_2$ norm can be added to the loss function, resulting in an overall loss function as: 
\begin{equation}
\begin{aligned}
& \mathcal{L}_{CE}= - \sum_{i=1}^N p_i \log q_i \; ,
\end{aligned}
\end{equation}
where $p_i$ and $q_i$ denote the ground-truth, and predicted probabilities for each image, respectively.
We can then minimize this loss function using stochastic gradient descent algorithm (and its variations).
We attempted to add regularization to the loss function, but the resulting model was not exhibiting a better performance.

\section{Experimental Results}
\label{sec_result}

%In this section we provide the experimental results of the four neural networks trained for COVID-19 detection, the histogram of their predicted scores on the test images, and  quantitative performance. %as well as the heatmap of the potentially infected regions in the Chest X-ray of images COVID-19 patients. 

\subsection{Model Hyper-parameters}
We fine-tuned each model for 100 epochs.
The batch size is set to 20, and ADAM optimizer is used to optimize the loss function, with a learning rate of 0.0001.
All images are down-sampled to 224x224 before being fed to the neural network (as these pre-trained models are usually trained with a specific image resolution).
All our implementations are done in PyTorch \citep{pytorch}, and are publicly available at https://github.com/shervinmin/DeepCovid.git

\subsection{Evaluation Metrics}
There are different metrics which can be used for evaluating the performance of classification models, such as classification accuracy, sensitivity, specificity, precision, and F1-score. 
Since the current test dataset is highly imbalanced (100  COVID-19 images, 3000 Non-COVID image), sensitivity and specificity are two proper metrics which can be used for reporting the model performance:
%. These metrics are also widely used in medical domain, and can be defined as Eq \ref{metrics}:
\begin{equation}
\begin{split}
\textbf{Sensitivity} & = \frac{\text{\#Images correctly predicted as COVID-19}}{\text{\#Total COVID-19 Images}} \; ,\\ 
\textbf{Specificity} & = \frac{\text{\#Images correctly predicted as Non-COVID}}{\text{\#Total Non-COVID Images}}\; .
\label{metrics}
\end{split}
\end{equation}

\subsection{Model Predicted Scores}
As mentioned earlier, we focused on four popular convolutional networks, ResNet18, ResNet50, SqueezeNet, DenseNet121.
These models predict a probability score for each image, which shows the likelihood of the image being detected as COVID-19.
By comparing this probability with a cut-off threshold, we can derive a binary label showing if the image is COVID-19 or not. 
An ideal model should predict the probability of all COVID-19 samples close to 1, and non-COVID samples close to 0.

\begin{figure}[h]
\begin{center}
    \includegraphics[width=0.8\linewidth]{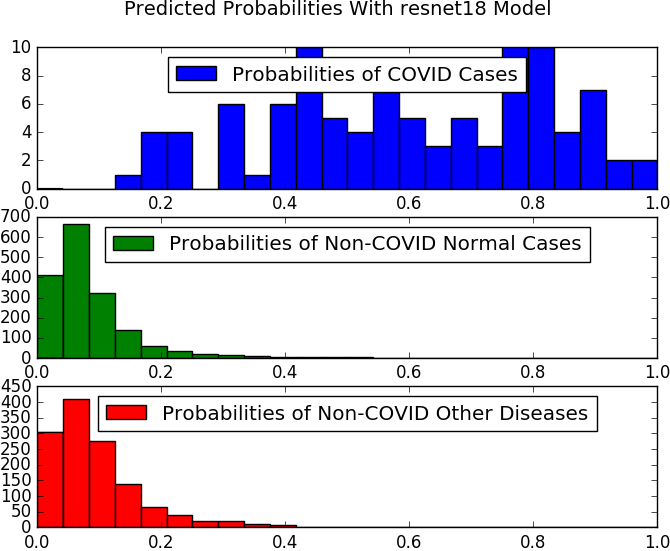}
\end{center}
  \caption{The predicted probability scores by ResNet18 on the test set.}
  \label{fig:resnet18_scores}
\end{figure}

\begin{figure}[h]
\begin{center}
    \includegraphics[width=0.8\linewidth]{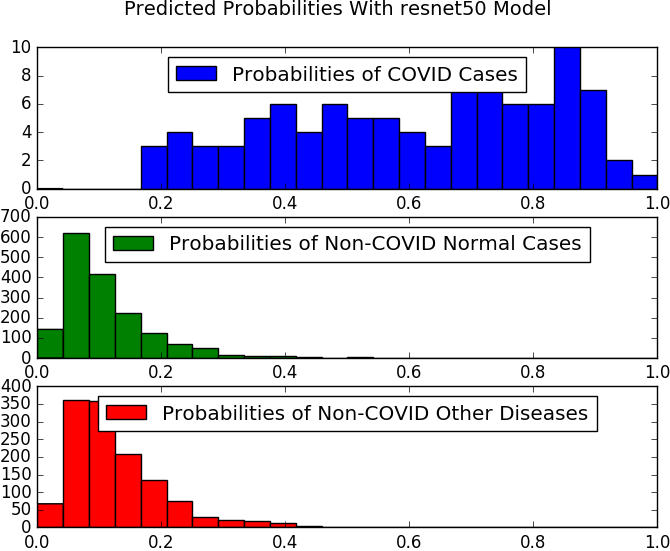}
\end{center}
  \caption{The predicted probability scores by ResNet50 on the test set.}
  \label{fig:resnet50_scores}
\end{figure}

\begin{figure}[h]
\begin{center}
    \includegraphics[width=0.8\linewidth]{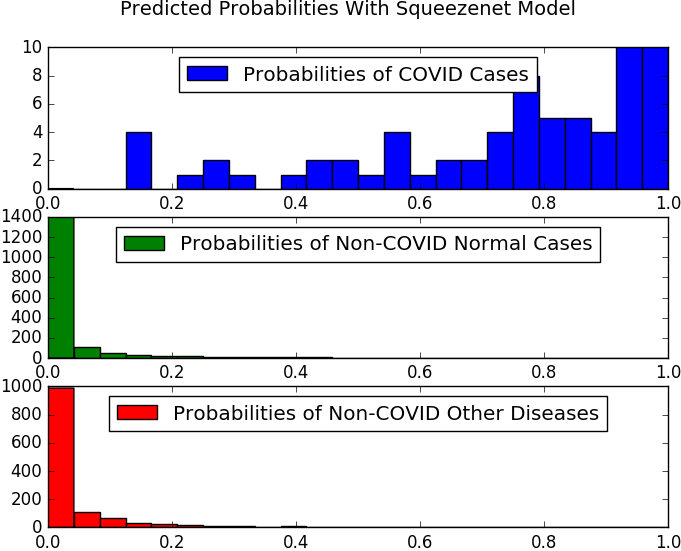}
\end{center}
  \caption{The predicted probability scores by SqueezeNet  on the test set.}
  \label{fig:SqueezeNet_scores}
\end{figure}

\begin{figure}[h]
\begin{center}
    \includegraphics[width=0.8\linewidth]{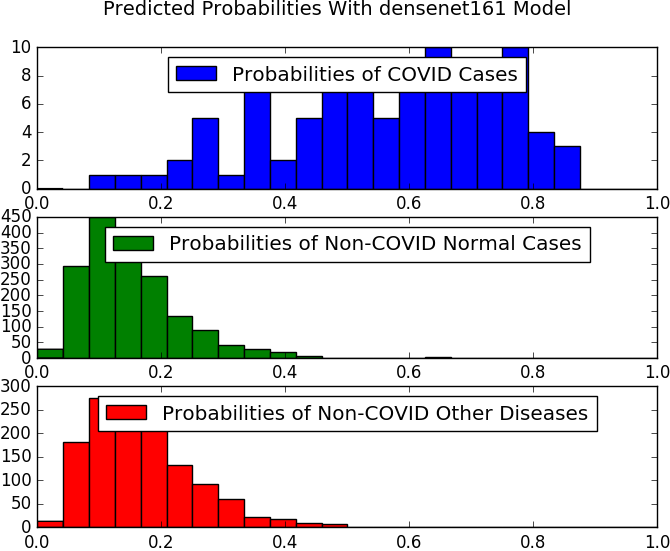}
\end{center}
  \caption{The predicted probability scores by DesneNet-121  on the test set.}
  \label{fig:densenet_scores}
\end{figure}

Figures \ref{fig:resnet18_scores},
\ref{fig:resnet50_scores},
\ref{fig:SqueezeNet_scores}, and \ref{fig:densenet_scores} show the distribution of predicted probability scores for the images in the test set, by ResNet18, ResNet50, SqueezeNet, and DenseNet-161 respectively.
%For each model, we show the predicted scores for Covid-19 test samples, non-Covid normal cases, as well as non-Covid cases which show other diseases (different from Covid-19).
Since Non-COVID class in our study contains both normal cases, as well as other types of diseases, we provide the distribution of predicted scores for three classes: COVID-19, Non-COVID normal, and Non-COVID other diseases.
As we can see the Non-Covid images with other  disease types have slightly larger scores than the Non-COVID normal cases.
This makes sense, since those images are more difficult to distinguish from COVID-19, than normal samples.

COVID-19 patient images are predicted to have much higher probabilities than the Non-COVID images, which is really encouraging, as it shows the model is learning to discriminate COVID-19 from non-COVID images.
Among different models, it can be observed that SqueezeNet does a much better job in pushing the predicted scores for COVID-19 and Non-COVID images farther apart from each other.

%%%% hereeee

\subsection{Model Sensitivity and Specificity}
Each model predicts a probability score showing the chance of the image being COVID-19. 
We can then compare these scores with a threshold to infer if the image is COVID-19 or not.
%(if the score is bigger than the threshold it will be predicted as COVID-19). 
The predicted labels are used to estimate the sensitivity and specificity of each model.
Depending on the value of the cut-off threshold, we can get different sensitivity and specificity rates for each model.

Tables \ref{tab:resnet_thresh}, \ref{tab:resnet50_thresh}, \ref{tab:squeezenet_thresh}, and \ref{tab:densenet_thresh} show the sensitivity and specificity rates for different thresholds, using ResNet18, ResNet50, SqueezeNet, and DenseNet-121 models, respectively.
As we can see, all these models achieve very promising results, and the best performing model obtains a sensitivity rate of 98\% and specificity rate of 92.9\%. 
%\xxx{MS: Earlier, when you summarized the accomplishments, you mentioned max specificity of 95\% - which one is true?}
SqueezeNet and ResNet18 achieve slightly better performance than the other models.
\begin{table}[ht]
\centering
  \caption{Sensitivity  and specificity rates of ResNet18 model, for different threshold values.}
  \centering
\begin{tabular}{|c|c|c|}
\hline
Threshold  & Sensitivity & Specificity\\
\hline
0.1  & 100\% & 72.4\% \\ \hline
0.17  & 98\% & 90.7\% \\ \hline
0.2  & 95\% & 92.4\% \\ \hline
0.25  & 91\% & 95.8\% \\ \hline
0.35 & 85\% & 98.3\% \\ \hline
\end{tabular}
\label{tab:resnet_thresh}
\end{table}

\begin{table}[ht]
\centering
  \caption{Sensitivity  and specificity rates of ResNet50 model, for different threshold values.}
  \centering
\begin{tabular}{|c|c|c|}
\hline
Threshold  & Sensitivity & Specificity\\
\hline
0.15  & 100\% & 78.2\% \\ \hline
0.205  & 98\% & 89.6\% \\ \hline
0.25  & 93\% & 94.2\% \\ \hline
0.3  & 90\% & 97.3\% \\ \hline
0.35  & 85\% & 98.4\% \\ \hline
\end{tabular}
\label{tab:resnet50_thresh}
\end{table}

\begin{table}[ht]
\centering
  \caption{Sensitivity  and specificity rates of SqueezeNet model, for different threshold values.}
  \centering
\begin{tabular}{|c|c|c|}
\hline
Threshold  & Sensitivity & Specificity\\
\hline
0.1  & 100\% & 89.9\% \\ \hline
0.15  & 98\% & 92.9\% \\ \hline
0.2  & 96.0\% & 94.6\% \\ \hline
0.4 & 92\% & 97.6\% \\ \hline
0.5 & 87\% & 98.3\% \\ \hline
\end{tabular}
\label{tab:squeezenet_thresh}
\end{table}

\begin{table}[ht]
\centering
  \caption{Sensitivity  and specificity rates of DenseNet-121 model, for different threshold values.}
  \centering
\begin{tabular}{|c|c|c|}
\hline
Threshold  & Sensitivity & Specificity\\
\hline
0.19  & 98\% & 75.1\% \\ \hline
0.25  & 95\% & 88.9\% \\ \hline
0.3 & 90\% & 94.6\% \\ \hline
0.4  & 79\% & 98.9\% \\ \hline
\end{tabular}
\label{tab:densenet_thresh}
\end{table}

\subsection{Small Number of COVID-19 Cases and Model Reliability}
It is worth mentioning that since so far the number of reliably labeled COVID-19 X-ray images is very limited, and we only have 100 test images in COVID-19 class, the sensitivity and specificity rates reported above may not be reliable. Ideally more experiments on a larger number of test samples with COVID-19 is needed to derive a more reliable estimation of sensitivity rates. 
We can however estimate the 95\% confidence interval of the reported sensitivity and specificity rates here, to see what is the possible range of these values for the current number of test samples in each class. 
The confidence interval of the accuracy rates can be calculated as:
%Eq.\ \ref{eq:confidence}:
\begin{equation}
\begin{aligned}
r= z \ \ \sqrt{\frac{accuracy \ (1-accuracy)}{N}} \; ,
\end{aligned}
\label{eq:confidence}
\end{equation}
where $z$ denotes the significance level  of the confidence interval (the number of standard deviation of the  Gaussian distribution), accuracy is the estimated accuracy (in our cases sensitivity and specificity), and $N$ denotes the number of samples for that class. 
Here we used 95\% confidence interval, for which the corresponding value of $z$ is 1.96.

As for COVID-19 diagnostic, having a sensitive model is crucial, we choose the cut-off threshold corresponding to a sensitivity rate of 98\% for each model, and compare their specificity rates.
Table \ref{table_comp} provides a comparison of the performance of these four models on the test set.
As we can see the confidence interval of specificity rates are small (around 1\%), since we have around 3000 samples for this class, whereas for the sensitivity rate we get slightly higher confidence interval (around 2.7\%) because of the limited number of samples.
\begin{table}[ht]
\centering
  \caption{Comparison of sensitivity and specificity of four state-of-the-art deep neural networks.}
  \centering
\begin{tabular}{|c|c|c|}
\hline
Model & Sensitivity & Specificity \\ \hline
ResNet18 & 98\% $\pm$ 2.7\% & 90.7\% $\pm$ 1.1\% \\ \hline
ResNet50  & 98\% $\pm$ 2.7\%& 89.6\% $\pm$ 1.1\% \\ \hline
SqueezeNet & 98\% $\pm$ 2.7\%& 92.9\% $\pm$ 0.9\% \\ \hline
Densenet-121  & 98\% $\pm$ 2.7\%& 75.1\% $\pm$ 1.5\% \\ \hline
\end{tabular}
\label{table_comp}
\end{table}

\subsection{The ROC Curve, Precision Recall Curve, and Confusion Matrix}
\begin{figure}[h]
\begin{center}
    \includegraphics[width=0.99\linewidth]{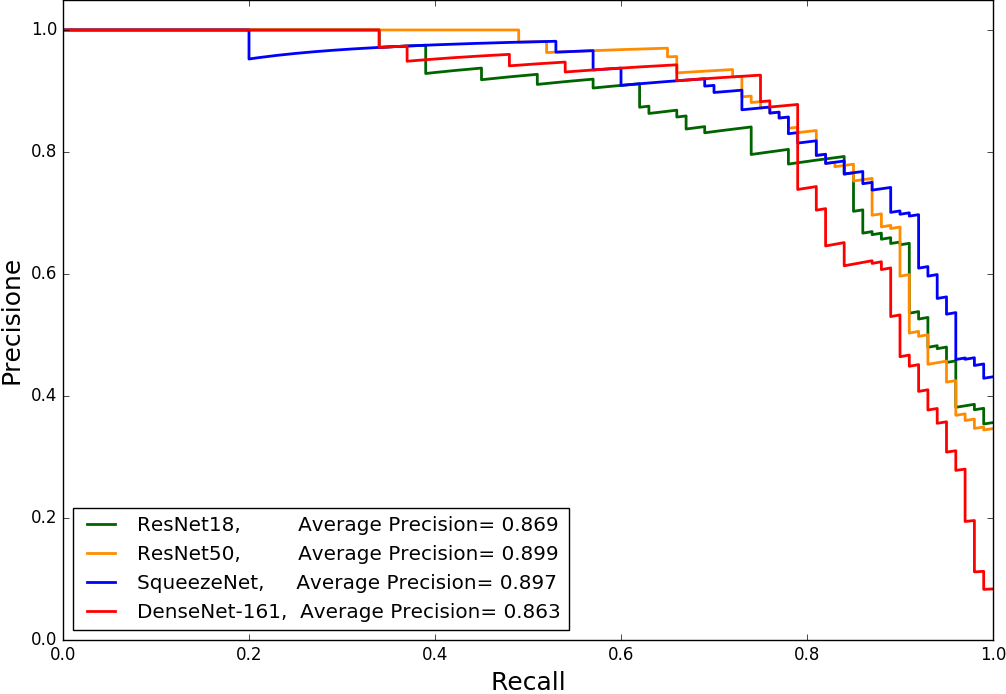}
\end{center}
  \caption{The precision-recall curve of four CNN architectures on test set.}
  \label{fig:prec_rec}
\end{figure}

It is hard to compare different models only based on their sensitivity and specificity rates, since these rates change by varying the cut-off thresholds. 
To see the overall comparison between these models, we need to look at the comparison for all possible threshold values.
One way to do this, is through the precision-recall curve, which provides the precision rate as a function of recall rate.
Precision is defined as the true positive images divided by the total number of images flagged as positive by the model, and the recall is the same as sensitivity rate (defined in Eq (2)).
The precision-recall curve of these four models is shown in Figure \ref{fig:prec_rec}.

\begin{figure}[h]
\begin{center}
    \includegraphics[width=0.99\linewidth]{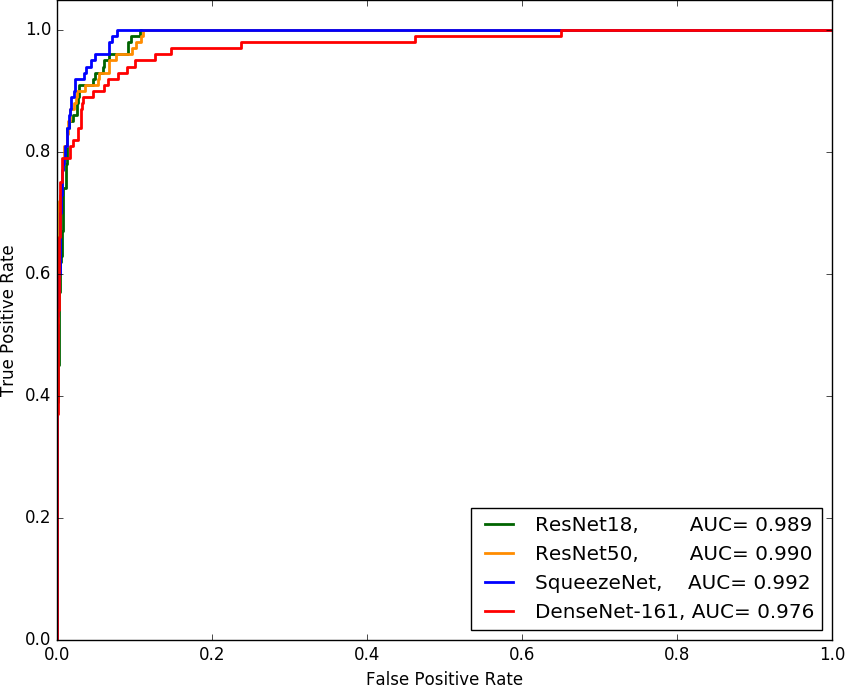}
\end{center}
  \caption{The ROC curve of four CNN architectures on COVID-19 test set.}
  \label{fig:roc}
\end{figure}

Another way to do this, is through the Receiver operating characteristic (ROC) curve, which provides the true positive rate as a function of false positive rate.
The ROC curve of these four models is shown in Figure \ref{fig:roc}. 
All models have a similar performance according to the AUC with the SqueezeNet achieving a slightly higher AUC than the other models.
It is worth mentioning that for highly imbalanced test sets, the AUC may not be a good indicative of model performance (as it can be very high), and looking at average-precision and precision-recall curve 
would be a better choice in that case. Here we provided both curves for the sake of completeness.

\begin{figure}[h]
\begin{center}
    \includegraphics[width=0.64\linewidth]{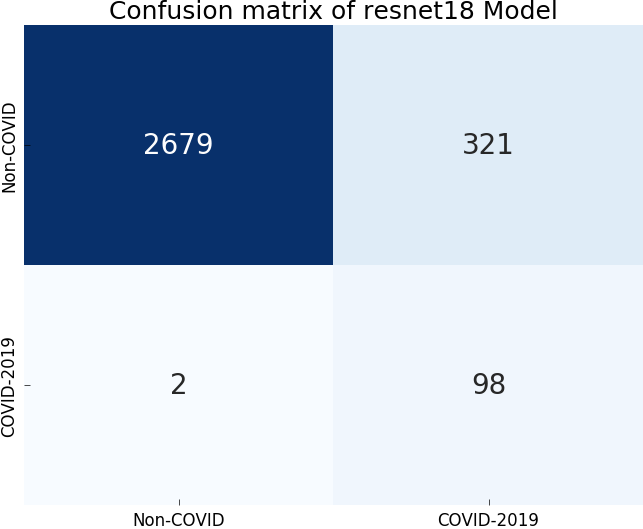}
\end{center}
  \caption{The confusion matrix of the proposed ResNet18 model.
  }
  \label{fig:conf_mat_res18}
\end{figure}

\begin{figure}[h]
\begin{center}
    \includegraphics[width=0.64\linewidth]{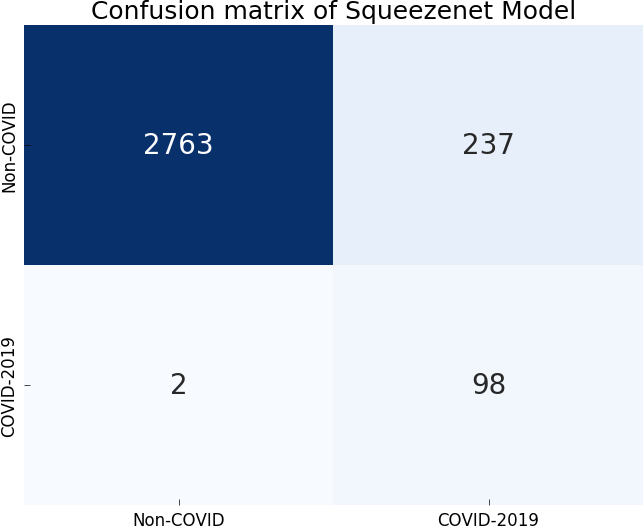}
\end{center}
  \caption{The confusion matrix of the proposed SqueezeNet framework.}
  \label{fig:conf_mat_squ}
\end{figure}

To see the exact number of correctly samples as COVID-19 and Non-COVID, the confusion matrices of the two top-performing models -- the fine-tuned ResNet18 and SqueezeNet -- when classifying the set of 3100 test images are shown in Figures  \ref{fig:conf_mat_res18} and \ref{fig:conf_mat_squ}.

\subsection{The Heatmap of Potentially Infected Regions} 
\begin{figure*}[h]
\begin{center}
    \includegraphics[width=0.99\linewidth]{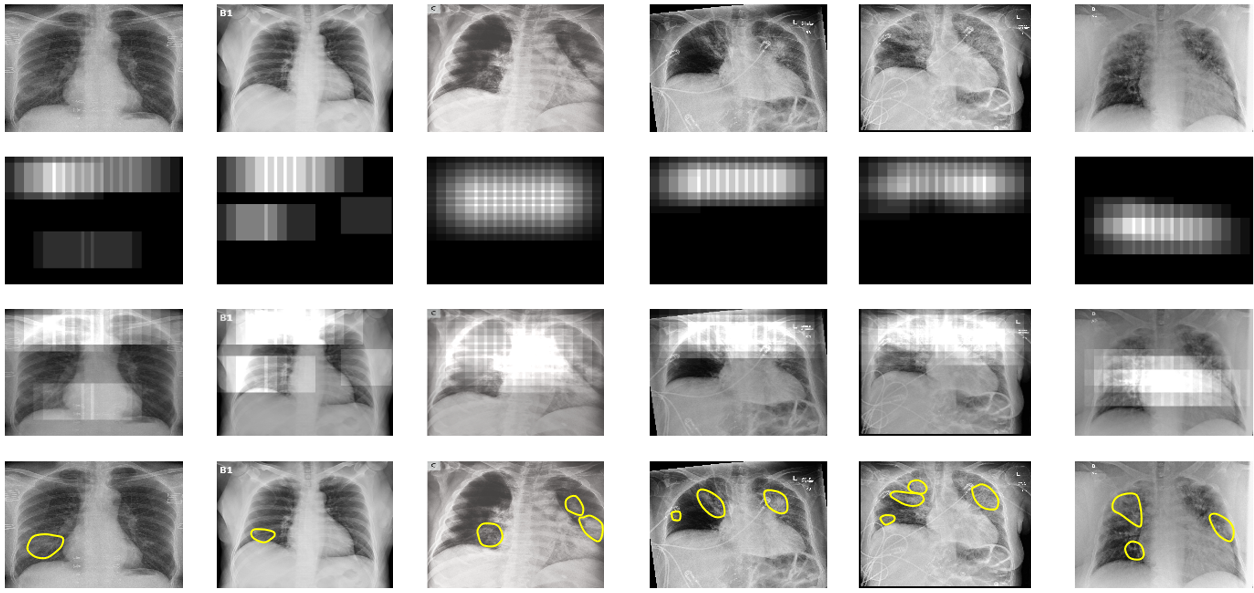}
\end{center}
  \caption{COVID-19 infected regions detected by our ResNet18 model, in six chest X-ray images from the test set. Vertical sets give the Original images (top row), COVID-19 region heatmap (2$^{nd}$ row), heatmap overlaid on the image (3rd row), and the independent standard of radiologist-marked COVID-19 disease regions (bottom row).
   %\xxx{MS: is it possible to make the radiologist markings better visible? Currently impossible to see unless you zoom-in a lot. We need to make this visible - otherwise we will not get the point across.-Should we make Fig.\ 13 much larger – maybe covering a full page – it seems way to small now.}
  }
\label{fig_heatmaps}
\end{figure*}

We used a simple technique to detect the potentially infected regions, while performing COVID-19 detection.
This technique is inspired by the work of Zeiler and Fergus \citep{fergus}, to visualize the result of deep convolutional networks.
We start from the top-left corner of the image, and each time occluding a square region of size $N$x$N$ inside the image, and make a prediction using the trained model on the occluded image.
If occluding that region causes the model to mis-classify a COVID-19 image as Non-COVID, that area would be considered as a potentially infected region in chest X-ray images (mainly because removing the information of that part led to model mis-classification).
On the other hand, if occluding a region does not impact the model's prediction, we infer that region is not infected.
Once we repeat this procedure for different sliding windows of $N$x$N$, each time shifting them with a stride of $S$, we can get a saliency map of the potentially infected regions in detecting COVID-19.
The detected regions for six example COVID-19 images from our test set are shown in Figure \ref{fig_heatmaps}.
The likely regions of COVID-19 disease marked by our board-certified radiologist are shown in blue on the last row.  The generated heatmaps show a good agreement with the radiologist-determined regions of the COVID-19 disease.

\section{Conclusion}
\label{sec_conc}

We reported a deep learning framework for COVID-19 detection from Chest X-ray images, by fine-tuning four pre-trained convolutional models (ResNet18, ResNet50, SqueezeNet, and DenseNet-121) on our training set.
We prepared a dataset of around 5k images, called COVID-Xray-5k (using images from two datasets), with the help of a board-certified radiologist to confirm the COVID-19 labels.
We make this dataset publicly available for the research community to use as a benchmark for training and evaluating future machine learning models for COVID-19 binary classification task.
We performed a detail experimental analysis evaluating the performance of each of these 4 models on the test set of of COVID-Xray-5k Dataset, in terms of sensitivity, specificity, ROC, and AUC.
For a sensitivity rate of 98\%, these models achieved a specificity rate of around 90\% on average. This is really encouraging, as it shows the promise of using X-ray images for COVID-19 diagnostics. 
This study is conducted on a set of publicly available images, which contains around 200 COVID-19 images, and 5,000 non-COVID images.
The presented work is reflecting one of the earliest Covid-19 chest X-ray analysis and dataset preparation attempts, which brings time-sensitive relevance in combining these two aspects.
However, due to the limited number of COVID-19 images publicly available so far,  further experiments are needed on a larger set of cleanly labeled COVID-19 images for a more reliable estimation of the accuracy of these models.
%The presented work is reflecting one of the earliest Covid-19 chest X-ray analysis and dataset preparation attempts, which brings time-sensitive relevance in combining these two aspects. There is no doubt that better and methodologically more in-depth contributions will follow by us and others. Yet, the timeliness of this work is of societal as well as data-sharing importance.

\section*{Acknowledgment}
The authors would like to thank Joseph Paul Cohen for collecting the COVID-Chestxray-dataset, and Sean Mullan for helping us with data preparation.
We would also like to thank the providers of ChexPert dataset, which are used as the negative samples in our case.
Milan Sonka’s research effort supported, in part, by NIH grant R01-EB004640.

%\section*{References}
%%Harvard
\bibliographystyle{IEEEtran}
%\biboptions{authoryear}
%\bibliography{refs}

\iffalse
\section*{Supplementary Material}

Supplementary material that may be helpful in the review process should
be prepared and provided as a separate electronic file. That file can
then be transformed into PDF format and submitted along with the
manuscript and graphic files to the appropriate editorial office.
\fi

\end{document}